\documentclass[runningheads]{llncs}

\usepackage{amsmath, graphicx, enumitem, multirow, amssymb}

\begin{document}

\title{Comparing 3D deformations between longitudinal daily CBCT acquisitions using CNN for head and neck radiotherapy toxicity prediction}

\titlerunning{Deformable CNN for HNC toxicity prediction on CBCT}

\author{
William Trung Le\inst{1}\orcidID{0000-0002-6106-7386} \and
Chulmin Bang\inst{2} \and
Philippine Cordelle\inst{1} \and
Daniel Markel\inst{2} \and
Phuc Felix Nguyen‐Tan\inst{2} \and
Houda Bahig\inst{2} \and
Samuel Kadoury\inst{1, 2}\orcidID{0000-0002-3048-4291}
}

\authorrunning{W. T. Le et al.}

\institute{Polytechnique Montr\'eal, Montreal, Quebec, Canada \and
Centre Hospitalier de l'Universit\'e de Montr\'eal, Montreal, Quebec, Canada
}

\maketitle

\begin{abstract}
Adaptive radiotherapy is a growing field of study in cancer treatment due to it's objective in sparing healthy tissue.
The standard of care in several institutions includes longitudinal cone-beam  computed tomography (CBCT) acquisitions to monitor changes, but have yet to be used to improve tumor control while managing side-effects.
The aim of this study is to demonstrate the clinical value of pre-treatment CBCT acquired daily during radiation therapy treatment for head and neck cancers for the downstream task of predicting severe toxicity occurrence: reactive feeding tube (NG), hospitalization and radionecrosis.
For this, we propose a deformable 3D classification pipeline that includes a component analyzing the Jacobian matrix of the deformation between planning CT and longitudinal CBCT, as well as clinical data.
The model is based on a multi-branch 3D residual convolutional neural network, while the CT to CBCT registration is based on a pair of VoxelMorph architectures.
Accuracies of $85.8\%$ and $75.3\%$ was found for radionecrosis and hospitalization, respectively, with similar performance as early as after the first week of treatment.
For NG tube risk, performance improves with increasing the timing of the CBCT fraction, reaching $83.1\%$ after the 5\textsuperscript{th} week of treatment.

\end{abstract}

\section{Introduction}
\label{sec:intro}

Radiation therapy has seen an increase in usage as a form of primary cancer treatment in recent years, due to significant improvements in imaging quality \cite{pan2016supply}.
However, radiotherapy is well known to be associated with negative side-effects.
Particularly in head and neck cancers (HNC), radiotherapy is associated with significant toxicities affecting the upper aerodigestive track causing impaired eating function or tissue necrosis \cite{radionecrosis}, leading to needing placement of a reactive feeding tube (NG) \cite{ngt} or even hospitalization \cite{hospitalization}.
To this day it remains an ongoing challenge to balance intensification vs de-intensification of dose delivery in order to simultaneously maximize tumor control and minimize affecting surrounding organs at risk (OAR).

In the past decade, it has become routine practice to include daily volumetric scans during treatment to better localize the targeted tumor and monitor anatomical changes.
While not as high quality as the computed tomography scan used for treatment planning (pCT), these serial or temporal cone-beam CTs (CBCT) acquired before each radiotherapy fraction contain a wealth of imaging biomarkers that are simply unused in the current standard of care.
Specifically, information on the deformation of the tissues and organs during therapy due to mass reduction or tumor shrinkage can be potentially associated with tumor control and toxicity outcomes.
Dysphagia for example is often seen correlated with anatomical changes in the surrounding areas of the digestive track \cite{ngt}.
This information could then potentially be used to adapt the radiotherapy plan (ART) in progress to minimize the risk of toxicities.
Recently, Muelas-Soria et al. \cite{muelas2022usefulness} found it possible to determine the CBCT fraction at which dosimetry re-planning would be required using a commercial deformable image registration (DIR) platform (RayStation).
Nonetheless, to our knowledge no work yet has tried to use ART based on patient's daily anatomical changes observed during treatment to predictive toxicity risks.

Many of the leading state-of-the-art methods in medical image analysis involve the use of deep neural networks (DNN) for efficient extraction of automatic features and task-specific therapy biomarkers.
Bibault et al. \cite{bibault2018deep} used a DNN method on radiomics features to evaluate treatment response for locally advanced rectal cancer.
However, methods leveraging convolutional neural networks (CNN) have been shown to be particularly efficient at processing large-scale 3D and 4D datasets, commonly used in radiology for diagnosis.
Jin et al. \cite{jin2021predicting} used a pair of Siamese subnetworks to evaluate treatment response which allowed analysis of dynamic information in longitudinal images, with pre- and post-op scans specifically.
Men et al. \cite{men2019deep} used a residual CNN to process pCT, organ segmentations and 3D dose distributions in order to predict xerostomia risk in patients with HNC.
FDG-PET/CT images were analyzed in Diamant et al. \cite{diamant2019} and Want et al. \cite{wang2020dose} using CNN architectures, the later including dose distribution in the inputs.
Recently, Le et al. \cite{presanet} adapted these techniques using a pseudo-3D version of the residual CNN \cite{resnet}, while also integrating patient clinical data as additional input features.

In this work, we design a HNC toxicity outcome prediction pipeline that analyses features obtained from the daily CBCTs, the deformations between those CBCT and the baseline pCT, and the patient's clinical records.
The proposed model is based on a multi-branch 3D version of the ResNet architecture \cite{resnet} processing 3D deformations along with an additional fully-connected branch for processing clinical data, with latent space fusion using concatenation before the final decision layers for binary output classification.
To encode the temporal aspect of a patient's response to treatment, a sub-step consisting of a pair of deep DIR model using the VoxelMorph framework was used to capture both the transformation from Housfield to CBCT space as well as the anatomical changes between the target CBCT\textsubscript{t} fraction the the pCT, transformed into a Jacobian matrix to capture the magnitude of the displacements.
We aim to determine the feasibility of predicting from radiotherapy-induced anatomical changes the risk of observing the following toxicities during treatment: reactive NG tube, hospitalization and radionecrosis.

The novel contributions proposed in this work are as follows:

\begin{enumerate}
  \item Analysis of CBCT images, which are yet clinically unproven for prognosis and ART.
  \item Classification of severe HNC radio-toxicities using the Jacobian matrix encoding of anatomical changes, learned using deep DIR models.
  \item Evaluation of the toxicity risk evolution based on CBCT fractions to maximize mid-treatment intervention opportunities.
\end{enumerate}

\section{Materials and Methods}
\label{sec:methods}

\subsection{Dataset and preprocessing}
\label{sec:dataset}

A retrospective dataset of 292 patients of the Centre Hospitalier de l'Universit\'e de Montr\'eal that underwent radiation therapy between 2000 and 2018 was used, with histologically confirmed HNC with or without surgery and concurrent chemotherapy.
This dataset includes 120kVp 1.5mm resolution pCT along with the associated OAR and tumor segmentation, between 30 to 35 daily pre-treatment 120kVp CBCT with 2mm resolution, and clinical data: age at diagnosis, sex, KPS, tumor location, smoker, alcohol consumption, cancer severity in the form of TNM staging \cite{tnm} (composed of primary tumor extent, regional lymph involvement, distant metastases), p16 biomarker status, whether the patient had a surgical tumor resection, and if the patient is undergoing concurrent chemotherapy.

Patient clinical data was one-hot encoded.
Tumor location includes: oropharynx, larynx, nasopharynx, hypopharynx, oral cavity or unknown primary.
All patients requiring a feeding tube at onset of therapy were excluded from the study.

All images were resampled isotropically to $2 \times 2 \times 2$mm and normalized to have pixel intensities between 0 and 1.
CT scans were cropped around the tumor segmentation (GTV) to a fix size of $128 \times 128 \times 128$ to match the field of view of the CBCTs, which were also center-cropped to the same shape.

\subsection{Registration and classification models}
\label{sec:models}

\begin{figure*}[ht]
    \begin{minipage}[b]{\linewidth}
      \centering
      \centerline{\includegraphics[width=\linewidth]{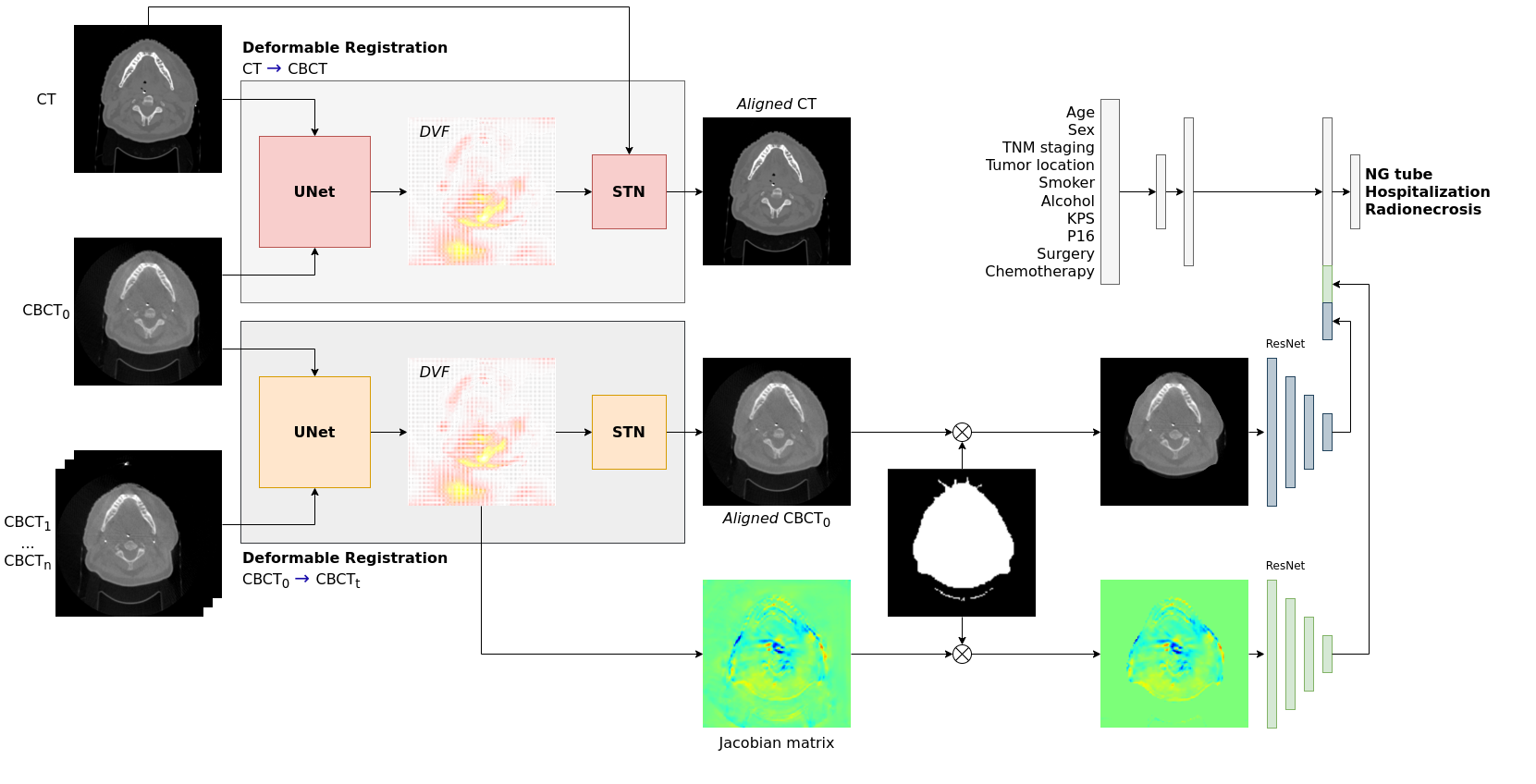}}
      %\centerline{(a) Result 1}\medskip
    \end{minipage}
    \caption{
        Multi-branch toxicity prediction model based on ResNet \cite{resnet}, adapted to 3D multi-modal inputs for binary classification.
        Two branch types were defined with their own sets of weights: a CBCT feature extractor, and an anatomical deformation analysis branch using the Jacobian matrix of the DVF between pCT and the target CBCT\textsubscript{t}.
        Masking was used in all images to include only the HNC region.
        A third fully connected input branch processed clinical data.
        Top left: deep DIR model trained to learn the CT $\rightarrow$ CBCT\textsubscript{0} mapping.
        Bottom left: deep DIR model trained to learnt the CBCT\textsubscript{0} $\rightarrow$ CBCT\textsubscript{t} mapping.
        Right: toxicity outcome classification model.
        Architecture shown is for the training phase.
        During testing, CBCT\textsubscript{0} is unused, and is replaced by the aligned CT.}
    \label{fig:classification}
\end{figure*}

Our proposed prediction pipeline contains an initial deep DIR offline training phase, followed by a second online classification phase.

In the initial step, the anatomical changes of the patient in response to radiotherapy over time were computed, that is between pCT and CBCT\textsubscript{t}.
While based on related technologies and seemingly similar visually, CT and CBCT present distinct characteristics in pixel intensities, edge resolution and artifacting \cite{cdfregnet}.
Older methods of DIR incorporated an initial intensity correction preprocessing step so that the registration could be seen as a mono-modal task \cite{twostage1,twostage2,twostage3}.
More recent approaches leveraging deep learning attempted to perform both tasks in a single fully automated way \cite{cdfregnet}.
Nonetheless, we are proposing to follow the former two stage approach precisely to distinguish between multi-modal mapping space from mono-modal transformations resulting uniquely from morphological changes in response to radiotherapy.
Thus by replacing the preprocessing steps with a more modern deep learning-based DIR approaches, we are able to train a pair of models based on the VoxelMorph \cite{voxelmorph} framework that are capable of disentangling the different stages of multi-modal registration.
And this will become relevant as our proposed method focuses specifically on extracting the transformation related to the second stage: registration resulting from anatomical changes before and during therapy.

To train under this proposed methodology, an initial model learned the mapping from pCT and CBCT\textsubscript{0} (see Figure \ref{fig:classification}).
Independently, a second model was trained to register CBCT that were deformed due to changes in anatomy resulting from radiotherapy treatment CBCT\textsubscript{0} $\rightarrow$ CBCT\textsubscript{t}).
Fixing the input image to CBCT\textsubscript{0} has the advantage of resembling the pCT the most since no treatment occurred between those two dates, which will become relevant at testing time.
At that stage then after both models are trained, the pCT is registered first to CBCT\textsubscript{0}, then that image (Aligned CT in Figure \ref{fig:classification}) is then fed into the second model and registered to the CBCT at the current fraction of treatment (not shown).

Both models used the UNet architecture \cite{unet} with $(16, 32, 32,32)$ layers in the encoder and $(32, 32, 32, 32, 32, 16, 16)$ layers in the decoder to generate the DVF outputs, as well as a spatial transformer network (STN) \cite{stn} to learn the deformation from the input to output space, as detailed in the original work \cite{voxelmorph}.
Due to the massive memory requirement for storing DVF, the final pCT $\rightarrow$ CBCT\textsubscript{t} deformations were converted to its Jacobian matrix $\mathbb{J}_f$
 which consists of the partial derivative linear approximation along each displacement vector, as in Eq. \ref{eq:jacobian}:

\begin{equation}
    \mathbb{J}_f = \left[\begin{array}{ccc}
        \dfrac{\partial \mathbf{f}(\mathbf{x})}{\partial x_{1}} &
        \cdots &
        \dfrac{\partial \mathbf{f}(\mathbf{x})}{\partial x_{n}}
        \end{array}\right]
    \label{eq:jacobian}
\end{equation}

where the transformation function $f$ corresponds to the composition of the two DIR models, as in Eq. \ref{eq:transformation}.
 
\begin{equation}
    \mathbf{f}_t: pCT \rightarrow CBCT_0 \rightarrow CBCT_t
    \label{eq:transformation}
\end{equation}

The multi-branch classification model itself is the 3D ResNet architecture --- using both bottleneck convolutional blocks and residual skip connections --- specifically the 34 and 50 layers variants as detailed in the original work by He et. al \cite{resnet} and provided by the MONAI library \cite{monai}.
To integrate features across the multiple inputs, separate branches were replicated without weights sharing to process the CBCT and $\mathbb{J}_f$ images.
These were then masked to remove non anatomical artifacts using adaptive thresholding.
An additional fully-connected branch with 3 layers (64, 128, 256) and ReLU activation integrated patient clinical data before fusion with the latent embedding vector of the image features.
Dropout of 40\% and batch normalization was used in every block of the clinical features path.

The final decision layer consisting of a single softmaxed fully-connected layer taking the concatenation of the CBCT, $\mathbb{J}_f$ and clinical data latent vectors produced a binary classification output.
A different model was trained for each event of reactive NG tube, hospitalization and radionecrosis outcome appearing at any point during the $30$\textendash$35$ days of treatment ($6$\textendash$7$ weeks of 5 work days).

\subsection{Training and evaluation}
\label{sec:training}

The sequential deep DIR models were trained offline to preprocess pCT/CBCT\textsubscript{t} pairs into a $\mathbb{J}_f$ matrix using the VoxelMorph framework \cite{voxelmorph}.
Inputs were first co-registered rigidly using the SimpleElastix software \cite{simpleelastix} with normalized cross-correlation loss (NCC), AdaptiveSGD optimizer and 4 levels of resolutions.
DIR models were both trained using the Adam optimizer with the global NCC image reconstruction loss and regularized by the L2 gradient loss.
In the first case, input pairs consisted of pCT/CBCT\textsubscript{0} with an L2 gradient regularization term of $\lambda = 1.0$.
In the second case, input pairs consisted of CBCT\textsubscript{0} and all available CBCT\textsubscript{t} with $t = {1 ... 35}$ with $\lambda = 0.5$.
In both cases, the registration models were trained with a $70\%$, $15\%$, $15\%$ split for training, validation and testing sets on paired input images only.
To quantitatively evaluate the performance of the end-to-end registration model, target registration errors based on 3 manually annotated landmarks was computed between the pCT and the final (30\textsuperscript{th}) CBCT fraction.

In the second online phase, a different model was trained for each of the 3 possible outcomes using a 5-fold cross-validation scheme.
Each model was trained to predict a binary target using the cross-entropy loss, weighted by the inverse of the toxicity frequency to combat data imbalance: the toxicity occurrence rate in the training dataset are $19.9\%$ for reactive NG tube, $3.8\%$ for radionecrosis and $7.2\%$ for hospitalization.
In the specific sub-study on the impact of multi-modal image inputs, the feature extractor branches used the hyperparameters defined by ResNet34, to account for the large increase in GPU memory requirements.
Otherwise, the ResNet50 architecture was used throughout this study \cite{resnet}.
In both the offline and online models, the OneCycle learning scheduling policy \cite{onecycle} was used to improve convergence time.

\section{Results}
\label{sec:results}

\begin{figure*}[htb]
    \centering
    \begin{minipage}[b]{0.135\linewidth}
      \centering
      \centerline{\includegraphics[width=\linewidth]{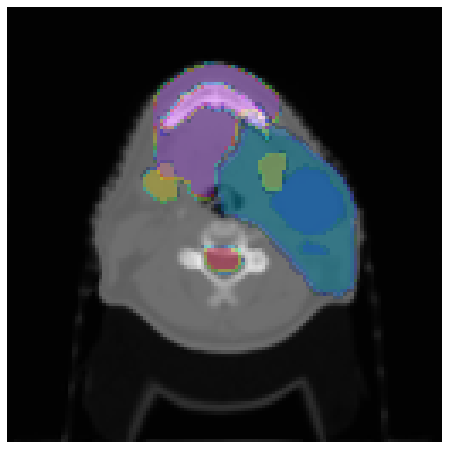}}
    \end{minipage}
    \begin{minipage}[b]{0.135\linewidth}
      \centering
      \centerline{\includegraphics[width=\linewidth]{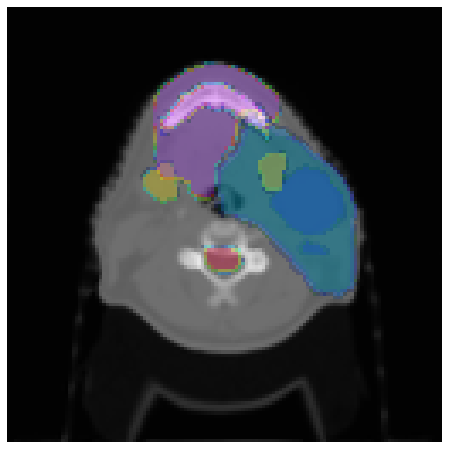}}
    \end{minipage}
    \begin{minipage}[b]{0.135\linewidth}
      \centering
      \centerline{\includegraphics[width=\linewidth]{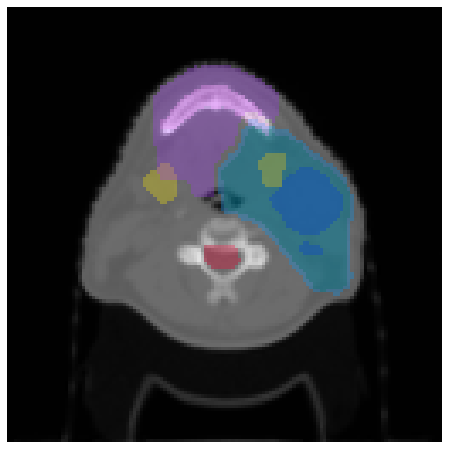}}
    \end{minipage}
    \begin{minipage}[b]{0.135\linewidth}
      \centering
      \centerline{\includegraphics[width=\linewidth]{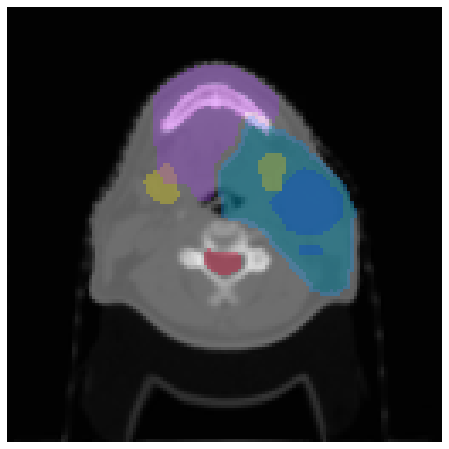}}
    \end{minipage}
    \begin{minipage}[b]{0.135\linewidth}
      \centering
      \centerline{\includegraphics[width=\linewidth]{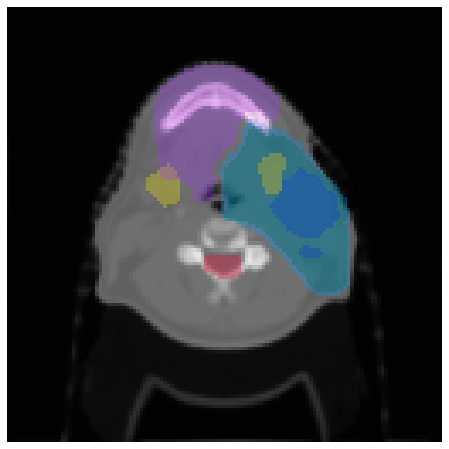}}
    \end{minipage}
    \begin{minipage}[b]{0.135\linewidth}
      \centering
      \centerline{\includegraphics[width=\linewidth]{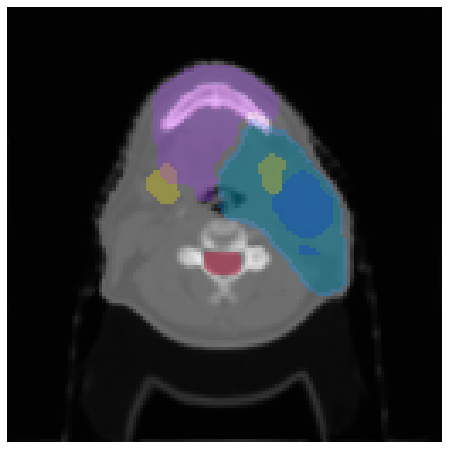}}
    \end{minipage}
    \begin{minipage}[b]{0.135\linewidth}
      \centering
      \centerline{\includegraphics[width=\linewidth]{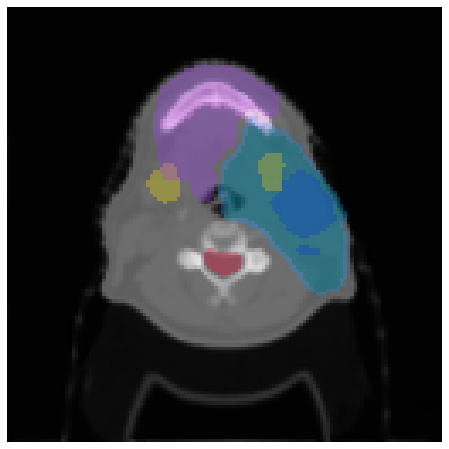}}
    \end{minipage}
    \begin{minipage}[b]{0.135\linewidth}
      \centering
      \centerline{\includegraphics[width=\linewidth]{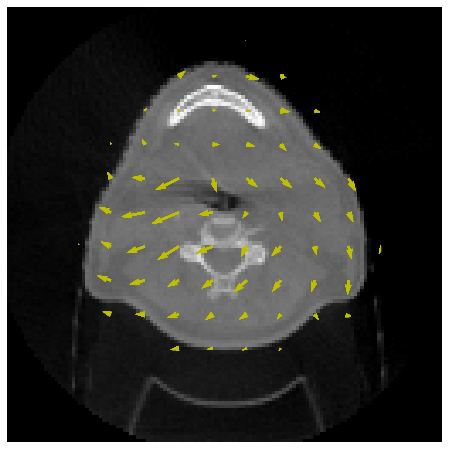}}
      \centerline{(a) CBCT\textsubscript{0}}\medskip
    \end{minipage}
    \begin{minipage}[b]{0.135\linewidth}
      \centering
      \centerline{\includegraphics[width=\linewidth]{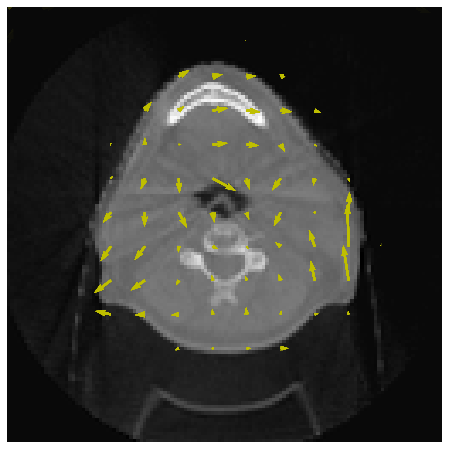}}
      \centerline{(b) CBCT\textsubscript{5}}\medskip
    \end{minipage}
    \begin{minipage}[b]{0.135\linewidth}
      \centering
      \centerline{\includegraphics[width=\linewidth]{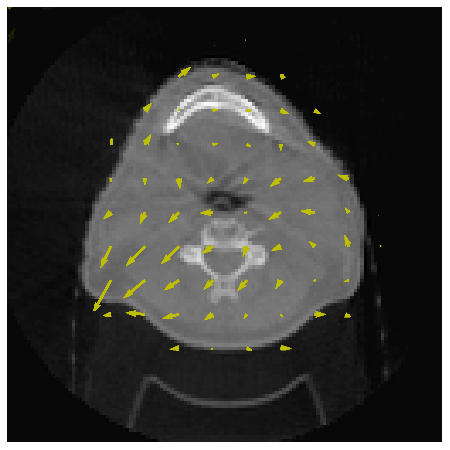}}
      \centerline{(c) CBCT\textsubscript{10}}\medskip
    \end{minipage}
    \begin{minipage}[b]{0.135\linewidth}
      \centering
      \centerline{\includegraphics[width=\linewidth]{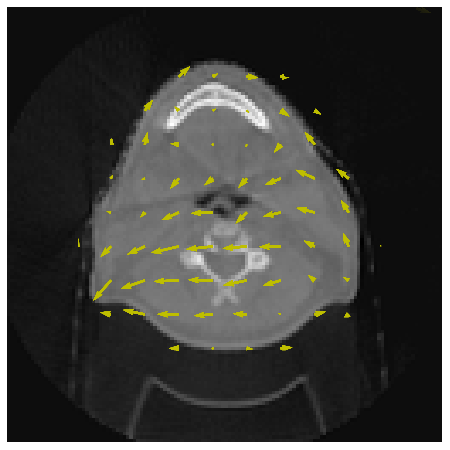}}
      \centerline{(d) CBCT\textsubscript{15}}\medskip
    \end{minipage}
    \begin{minipage}[b]{0.135\linewidth}
      \centering
      \centerline{\includegraphics[width=\linewidth]{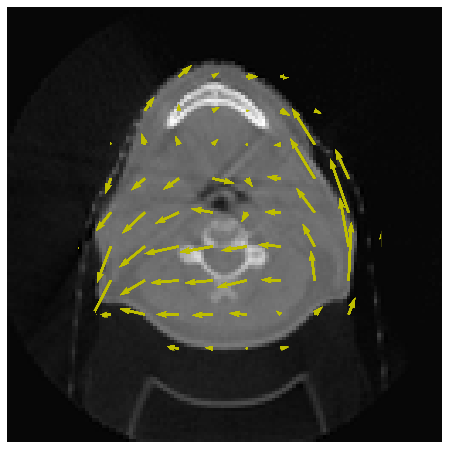}}
      \centerline{(e) CBCT\textsubscript{20}}\medskip
    \end{minipage}
    \begin{minipage}[b]{0.135\linewidth}
      \centering
      \centerline{\includegraphics[width=\linewidth]{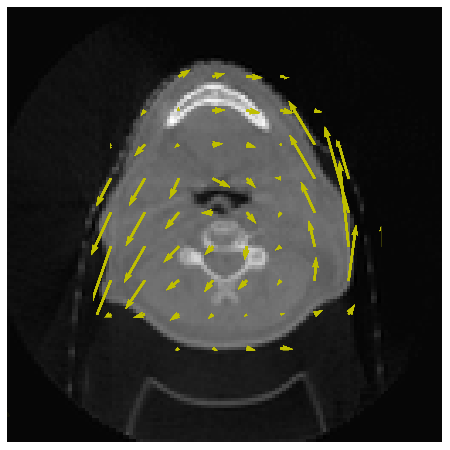}}
      \centerline{(f) CBCT\textsubscript{25}}\medskip
    \end{minipage}
    \begin{minipage}[b]{0.135\linewidth}
      \centering
      \centerline{\includegraphics[width=\linewidth]{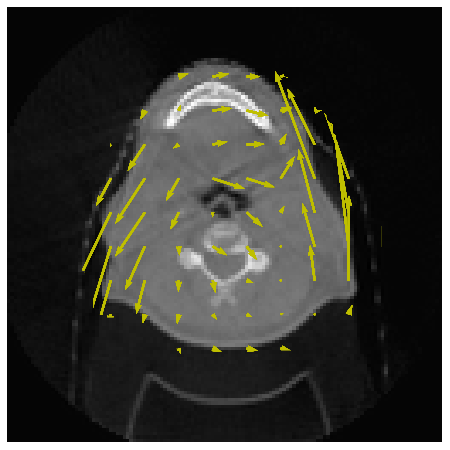}}
      \centerline{(g) CBCT\textsubscript{30}}\medskip
    \end{minipage}
    \caption{
        Registration of pCT to CBCT at different fractions with corresponding deformed OAR and tumor segmentation.
        Top: deformed pCT to the corresponding CBCT.
        The following segmentations are visible: oral cavity in pink, submandibular glands in yellow, GTV in blue and PTV in green.
        Bottom: Reference CBCT with DVF as yellow arrows.}
    \label{fig:qualitative}
\end{figure*}

The multi-modal registration model mapping CT inputs to CBCT voxel space was trained for for 100 epochs with early stopping with a batch size of 7.
An average validation loss of $-0.9438$ was obtained, with $99.9\%$ of the $\mathbb{J}_f$ matrix showing a non-zero deformation amplitude.
For the temporal CBCT DIR model, a $-0.9862$ mean validation NCC was obtained with deformations present in $99.4\%$ of the $\mathbb{J}_f$.
Qualitative results of the deformed pCT and segmentations are shown in Figure \ref{fig:qualitative} along with the reference CBCT fraction.
Mean TRE for landmarks yielded $2.83mm \pm 0.89$ on the 2D slice with visibly segmented tumor.

The classification model was trained for 100 epochs, with a batch size of 8 and the Adam optimizer with 0.0007 learning rate, except for the $\mathbb{J}_f$/clinical ablation study, which was trained for 200 epochs.
Performance metrics reported in the experiments include balanced accuracy (bAcc), specificity (Spec) and sensitivity (Sens).
To evaluate the correlation between the CBCT fraction and classification performance, a linear fit was used with a statistical significance of $r^2 \geq 0.9$.
All models were trained on Nvidia GeForce RTX 2080 Ti GPUs.

\begin{table}[ht]
    \centering
    \scalebox{1}{
    \begin{tabular}{c c c c c c}
         Toxicity & Clinical & $\mathbb{J}_f$ &  bAcc (\% $\pm$ std) & Spec (\% $\pm$ std) & Sens (\% $\pm$ std) \\
         \hline\hline
         \multirow{3}{*}{NG tube} & \checkmark & \checkmark & \textbf{69.1 $\pm$ 4.5} & \textbf{75.9 $\pm$ 18.4} & 61.7 $\pm$ 16.1 \\
         & & \checkmark & 68.8 $\pm$ 6.3 & 75.8 $\pm$ 21.8 & 61.8 $\pm$ 17.8 \\
         & \checkmark & & 60.0 $\pm$ 5.7 & 55.4 $\pm$ 12.2 & \textbf{64.6 $\pm$ 17.5} \\
         \hline
         \multirow{3}{*}{Hospitalization} & \checkmark & \checkmark & \textbf{75.3 $\pm$ 3.6} & 67.4 $\pm$ 16.6 & \textbf{83.1 $\pm$ 10.4} \\
         & & \checkmark & 72.9 $\pm$ 2.7 & \textbf{74.6 $\pm$ 12.0} & 71.1 $\pm$ 11.1 \\
         & \checkmark & & 56.4 $\pm$ 12.2 & 63.6 $\pm$ 21.2 & 49.3 $\pm$ 36.7 \\
         \hline
         \multirow{3}{*}{Radionecrosis} & \checkmark & \checkmark & \textbf{85.8 $\pm$ 7.7} & 78.2 $\pm$ 15.6 & \textbf{93.3 $\pm$ 14.9} \\
         & & \checkmark & 74.3 $\pm$ 10.4 & \textbf{93.3 $\pm$ 14.9} & 55.3 $\pm$ 27.7 \\
         & \checkmark & & 81.0 $\pm$ 14.9 & 75.4 $\pm$ 4.1 & 86.7 $\pm$ 29.8 \\
    \end{tabular}
    }
    \caption{
        Ablation study of the classification performance for models trained against $\mathbb{J}_f$ and clinical inputs.
        The baseline model contains only the feed-forward path for classifying clinical data.
        Models with deformation inputs in the form of the $\mathbb{J}_f$ matrix between the baseline CT and the 10\textsuperscript{th} day CBCT (chosen to balance between amount of deformation from baseline and distance from the end of treatment) add an additional ResNet50-like\cite{resnet} 3D convolutional branch before the classification layers.
        Figures are mean metrics over a 5-fold cross-validation with standard deviation.}
    \label{tab:benchmark}
\end{table}

Table \ref{tab:benchmark} shows the performance of the ablation study evaluating the effects of using anatomical deformation and patient clinical data as inputs to the model.
Results show that a combination of inputs leads to an improvement in accuracy vs baseline for each of reactive NG tube ($69.1\% > 60.0\%$), hospitalization ($75.3\% > 56.4\%$) and radionecrosis ($85.8\% > 81.0\%$).
Sensitivity was higher than specificity for hospitalization ($83.1\% > 67.4\%$) and radionecrosis ($93.3\% > 78.2\%$), but lower for reactive NG tube ($61.7\% < 75.9\%$).
Notably, reactive NG tube also has the lowest classification performance across all models.

\begin{table}[ht]
    \centering
    \scalebox{1}{
    \begin{tabular}{c c c c c c}
         Toxicity & CBCT & $\mathbb{J}_f$ & bAcc (\% $\pm$ std) & Spec (\% $\pm$ std) & Sens (\% $\pm$ std) \\
         \hline\hline
         \multirow{3}{*}{NG tube} & \checkmark & \checkmark & 64.7 $\pm$ 5.3 & 52.1 $\pm$ 13.0 & \textbf{77.2 $\pm$ 9.8} \\
         & & \checkmark & 68.2 $\pm$ 4.8 & 66.6 $\pm$ 11.2 & 69.8 $\pm$ 13.7 \\
         & \checkmark & & 68.2 $\pm$ 6.7 & \textbf{78.1 $\pm$ 13.0} & 58.3 $\pm$ 20.5 \\
         \hline
         \multirow{3}{*}{Hospitalization} & \checkmark & \checkmark & \textbf{79.6 $\pm$ 6.7} & \textbf{83.0 $\pm$ 5.4} & \textbf{76.3 $\pm$ 14.6} \\
         & & \checkmark & 64.4 $\pm$ 3.5 & 72.2 $\pm$ 15.5 & 56.6 $\pm$ 14.4 \\
         & \checkmark & & 73.6 $\pm$ 4.6 & 75.9 $\pm$ 6.2 & 71.3 $\pm$ 12.4 \\
         \hline
         \multirow{3}{*}{Radionecrosis} & \checkmark & \checkmark & 79.0 $\pm$ 10.9 & 68.0  $\pm$ 26.4 & 90.0 $\pm$ 22.4 \\
         & & \checkmark & 74.7 $\pm$ 12.7 & \textbf{72.8 $\pm$ 18.4} & 76.7 $\pm$ 22.4 \\
         & \checkmark & & \textbf{79.7 $\pm$ 15.6} & 59.4 $\pm$ 31.2 & \textbf{100 $\pm$ 0.0} \\
    \end{tabular}
    }
    \caption{
        Ablation study of the classification performance for models trained against $\mathbb{J}_f$ and CBCT inputs.
        All models include a clinical data processing branch as in Table \ref{tab:benchmark}.
        Multi-branch classification model include a separate ResNet34-like\cite{resnet} 3D convolutional branch per input modality before concatenation for the final classification binary output.
        Figures are mean metrics over a 5-fold cross-validation with standard deviation\textit{.}}
    \label{tab:cbct}
\end{table}

Table \ref{tab:cbct} shows the effects on classification performance when adding either or both of the CBCTs and the $\mathbb{J}_f$ to the inputs.
Improvements were found by substituting $\mathbb{J}_f$ for the CBCT volume for hospitalization ($73.6\% > 64.4\%$) and radionecrosis ($79.7\% > 74.7\%$) prediction, with no false negative errors in the later case.
Further improvements were found when combining both $\mathbb{J}_f$ and CBCT inputs for hospitalization prediction ($79.6\% > 73.6\%$).

\begin{figure}[ht]
    \begin{minipage}[b]{\linewidth}
      \centering
      \centerline{\includegraphics[width=\linewidth]{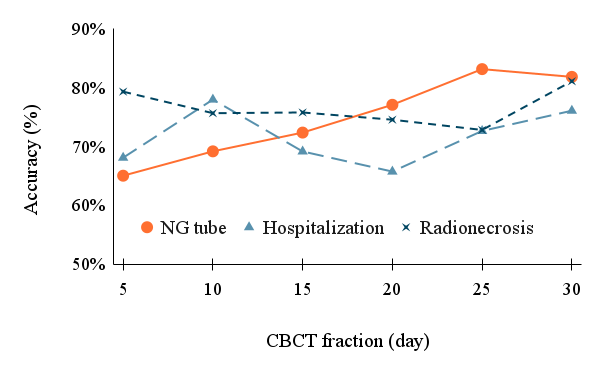}}
    \end{minipage}
    \caption{
        Risk evolution study.
        Balanced accuracy performance of predictive model trained against $\mathbb{J}_f$ resulting from deformations after patients received $t$ amount of radiotherapy fractions (day).
        Toxicity outcomes: $\circ$ NG tube, $\blacktriangle$ hospitalization, $\times$ radionecrosis.
        Dashed lines: no linear correlation found between CBCT fraction inputs and classification performance ($r^2 < 0.9$).
        Note that the baseline models with deformations at fraction day 0 corresponds to the model with clinical data only as inputs (not shown) as in Table \ref{tab:benchmark}.}
    \label{fig:evolution}
\end{figure}

Figure \ref{fig:evolution} shows the classification performance of the top-performing model with $\mathbb{J}_f$ and clinical data inputs as found in Table \ref{tab:benchmark} when trained against deformations at increasing fractions.
A positive correlation was found between the predictive power and the amount of radiation received for reactive NG tube classification ($r^2 > 0.9$), with a maximum accuracy of $83.1\%$ at the 25\textsuperscript{th} fraction.
No correlation was found for radionecrosis or hospitalization outcome prediction.

\section{Discussion}
\label{sec:discussion}

In this study, we demonstrated the clinical value of longitudinal pre-treatment CBCT for the downstream task of predicting radiotherapy toxicity risk in HNC.
The use of both the raw CBCT modality, despite not containing Hounsfield data \cite{mah2010deriving}, as well as the deformation between the target CBCT at the pCT, both allowed prediction of reactive NG tube, hospitalization and radionecrosis, all at the 10\textsuperscript{th} fraction.
However, performance was not as could be in the Jacobian + CBCT study, due to the lower model capacity used (ResNet-34 instead of ResNet-50 in the Jacobian-only model), a trade-off that was necessary arising from the memory limitation of the available GPU.

In addition, we showed that patient clinical data had a positive impact on model performance when combined with the imaging inputs ($4.8\%$ improvements for radionecrosis and $18.9\%$ for hospitalization), but lead to poor predictive power when used alone, as is the case of reactive NG tube prediction that barely performed better than random ($57.2\%$).

Furthermore, since predicting the risk of side effects loses clinical utility as the treatment progresses given the decreasing opportunity for intervention, we aimed to show whether there was a correlation between predictive power and the CBCT fraction used.
For reactive NG tube, performance improved over time, which suggested that the model was able to detect the link between a change in tissue mass and risk of dysphagia leading to requiring a feeding tube.
For radionecrosis and hospitalization however, no correlation was found, with all models showing similar performances between the 5\textsuperscript{th} and the 30\textsuperscript{th} treatment day.
A possibility is that these toxicities may be related to the tumor location and baseline radiological characteristics rather than the dynamic changes, which compels further examination.
Another would require correlating the time at which these two toxicities occurred in the patient cohort and the choice of fraction used as inputs to the models within the context of these longitudinal correlation experiments.
Moreover, given that the addition of CBCT images to the inputs seemed to improve overall classification scores for these two outcomes compared to baseline, additional investigation would be required to properly estimate the model's predictive ability that does not take the lower model capacity trade-off.
Finally, the main limitation the feel that needs to be addressed is the absence of use of radiotherapy specific imaging data such as tumor segmentation and dosimetry in combination with the available pre-treatment CBCT, the subject of a ongoing study.

\bibliographystyle{splncs04}
\bibliography{refs}

\end{document}